\documentclass[review]{elsarticle}

\usepackage{lineno,hyperref}
\usepackage{mathtools}
\usepackage{amsmath}
\modulolinenumbers[5]
\usepackage[symbol]{footmisc}

\begin{document}

\begin{frontmatter}

\title{Prediction and optimization of mechanical properties of composites using convolutional neural networks}



\author[daff]{Diab W. Abueidda}

\author[maff]{Mohammad Almasri}

\author[naff]{Rami ِAmmourah}

\author[maff]{Umberto Ravaioli}

\author[daff]{Iwona M. Jasiuk\corref{cor1}}
\cortext[cor1]{Corresponding authors}
\ead{ijasiuk@illinois.edu}

\author[naff]{Nahil A. Sobh\corref{cor1}}
\ead{sobh@illinois.edu}

\address[daff]{Department of Mechanical Science and Engineering, University of Illinois at Urbana-Champaign, 1206 West Green Street, Urbana, IL 61801-2906, USA}
\address[maff]{Department of Electrical and Computer Engineering, University of Illinois at Urbana-Champaign, 1308 West Main Street, Urbana, IL 61801, USA}
\address[naff]{Beckman Institute and Department of Civil and Environmental Engineering, University of Illinois at Urbana-Champaign, 205 North Mathews Avenue, Urbana, Illinois 61801, USA}

\begin{abstract}
In this paper, we develop a convolutional neural network model to predict the mechanical properties of a two-dimensional checkerboard composite quantitatively. The checkerboard composite possesses two phases, one phase is soft and ductile while the other is stiff and brittle. The ground-truth data used in the training process are obtained from finite element analyses under the assumption of plane stress. Monte Carlo simulations and central limit theorem are used to find the size of the dataset needed. Once the training process is completed, the developed model is validated using data unseen during training. The developed neural network model captures the stiffness, strength, and toughness of checkerboard composites with high accuracy. Also, we integrate the developed model with a genetic algorithm optimizer to identify the optimal microstructural designs. The genetic algorithm optimizer adopted here has several operators, selection, crossover, mutation, and elitism. The optimizer converges to configurations with highly enhanced properties. For the case of the modulus and starting from randomly-initialized generation, the GA optimizer converges to the global maximum which involves no soft elements. Also, the GA optimizers, when used to maximize strength and toughness, tend towards having soft elements in the region next to the crack tip. 
\end{abstract}


\end{frontmatter}

\section{Introduction} \label{intro_sec}
The pursuit of materials possessing robust properties has been of high scientific and industrial interests to meet the requirement of modern engineering applications necessitating advanced composite materials \cite{chawla2012composite, chen2017nacre, studart2013biological}. There are several approaches to manufacture composite materials. For instance, the widely used laminate composites are layers sequentially stacked to achieve the desired properties under predetermined loading conditions. The difficulty in achieving a strong adhesion between base materials has confined the laminate stacking process and limited the expedition to explore new composite materials. Also, composites have been made by mixing in fillers in a molten matrix. Another prominent approach is to rely on recent advances in additive manufacturing that enables the fabrication of complex combinations of distinct materials and tuning their properties in three spatial directions. With this freedom in fabricating composite materials, researchers have more flexibility to design materials with superior properties \cite{wang20173d,jasiuk2018overview}.  

Generally, one can develop new materials with desired properties through inspiration from natural and biological systems, optimization algorithms, or by combining both. Human bone is an example of a multifunctional material which achieves excellent mechanical properties due to its several distinct hierarchical levels \cite{reznikov2014bone,abueidda2017modeling,schwarcz2017ultrastructure}. Bone consists of a hard shell and soft core, where cortical bone (stiff) embraces trabecular bone (soft).  Bioinspiration and optimization provide superior material properties through the proper selection of the constituent materials and their volume fractions and identification of the optimal geometric configurations (spatial distributions of the constituents). One can combine the different approaches to optimize the performance of the materials further. In the past, researchers extensively studied the development of new materials by identifying the proper constituent materials and corresponding volume fraction, and such techniques are almost mature. Developed materials usually have taken the form of composites and cellular materials, and they yield properties not available in their bulk material counterparts \cite{hertzberg1996deformation,khaderi2014stiffness,chawla1998ceramic,al2018nature,tancogne20183d}. Although the foci of this paper are on the prediction and optimization of composite materials' properties, many analogies can be drawn to developing cellular materials with enhanced properties. Composite materials with randomly distributed constituents are prevalent and intensively studied by many researchers not only because of their rich physics and complexity compared to ones with deterministic distributions, but also because they are widely spread in nature  \cite{ostoja2007microstructural,khokhar2009simulations,dalaq2013scaling}. 

Employing brute force (also known as exhaustive search) algorithms to obtain optimized materials is not practical in most of the cases due to the enormous design space. Brute force algorithms are based on attempting all possible designs to identify composite materials with optimal microstructural material distributions. A more efficient approach is to use topology optimization algorithms to optimize mechanical properties. Most topology optimization algorithms available in the literature are gradient-based \cite{bendsoe2001topology,sigmund200199}. Both approaches mentioned above require solving many numerical simulations, and each simulation commonly takes from seconds to hours depending on the complexity of the problem. However, such techniques have an inherent limitation in terms of computational cost due to a large number of design variables and/or difficulties in finding the gradients in the case of gradient-based optimization. Using another platform which is faster than the finite element analysis and other available techniques such boundary element method to predict the mechanical properties of composite materials may revolutionize the field of composite materials' optimization.

Machine learning has been proven to be a potent tool in various applications \cite{michalski2013machine}. Machine learning is a statistical and predictive tool that helps to better perceive the behavior of a particular set of data and a problem. Recently, machine learning has been intensively used in spam detection, speech and image detection, search engines, and disease and drug discoveries \cite{witten2016data,lafon2006diffusion}. Applications of machine learning are not limited to the ones above; machine learning has been employed to predict the properties of different structural and material systems and to search for new materials with optimal designs \cite{agrawal2016perspective,goh2017deep,bisagni2002post,gu2018bioinspired}. 

Deep neural networks (DNNs) are one of the most powerful machine learning techniques. DNNs are inspired by architectures of biological neural networks simulating the way we humans learn from data. DNNs usually have an input layer, hidden layers, and an output layer. Several studies used DNNs to investigate material and structural behavior. Do et al. \cite{do2019material} used deep neural networks to supersede the finite element analysis while solving optimization problems (buckling and free vibration with various volume constraints) of functionally graded plates. In a different work, Nguyen et al. \cite{nguyen2018deep} used a deep neural network model for predicting the compressive strength of foamed concrete. 

Additionally, Kim et al. \cite{KIM20191} argued that deep learning networks could be employed to capture the nonlinear hysteretic systems without compromising accuracy. Although the adopted approach is generic and is suitable for any system with hysteresis, they applied it to the prediction of structural responses under earthquake excitations. Gopalakrishnan et al. \cite{gopalakrishnan2017deep} utilized deep neural network models trained on a large set of images, and then they transferred their learning capability to pavement crack detection using digitized pavement surface images. Such an approach is known as transfer learning in the field of machine learning. Additionally, Lee et al. \cite{lee2018background} used single-layered and multi-layered perception networks to study the well-known ten bar truss problem, and they investigated the effect of different hyper-parameters.

Recently, Gu et al. \cite{gu2018novo} used machine learning to study two mechanical properties (strength and toughness) of two-dimensional (2D) checkerboard composites possessing two constituent materials. The authors argued that machine learning is a very promising tool for investigating the mechanical properties of composite materials, and it can be incorporated in optimization algorithms to find designs with optimal performance. The authors developed two binary classifiers using a single layer convolutional neural network and a linear model to conclude whether a composite material with a specific material distribution yields good or bad mechanical properties. They found that a small proportion of the design space can be utilized to train the machine learning models that are capable of classifying the performance of the two-dimensional (2D) composites with high accuracy.

In this paper, we extend the work of Gu et al. \cite{gu2018novo}. Their sample space has 35960 composite configurations as they fixed the volume fractions of the phases; 12.5\% of the volume fraction is assigned to be a soft material, and the remaining 87.5\% is assigned as the stiff material. Also, the model they developed classifies a given composite as good or bad (qualitative prediction) rather than capturing the mechanical properties quantitatively. In this work, we develop a convolutional neural network model (CNN) that is capable of quantitatively predicting the mechanical properties of the composite over the entire volume fraction space. CNNs are a class of DNNs, and they are chosen because they have proven to be very successful in image recognition tasks. In our problem, we represent our checkerboard composites as images exploiting the robustness of CNNs. Accounting for all volume fractions dramatically enlarges the sample space as we discuss in the following sections. Also, from a practical point of view, it is more useful to have a model that predicts the performance of the composites in terms of real values rather than a set of classes obtained from a classification model. Moreover, we integrate an optimization scheme based on genetic algorithm with the developed CNN model to optimize the mechanical properties with respect to the volume fractions of the stiff and soft materials and their spatial distribution in the microstructure. 

\section{Methodology}

A convolutional neural network model is developed to quantitatively predict the stiffness, strength, and toughness of a 2D checkerboard composite system composed of two materials, one is soft and ductile, and the other is stiff and brittle. CNN models usually need a significant amount of data to provide accurate predictions. After training the CNN model, it is tested against new data unseen by the model during the training process. Our CNN model is trained on data generated from a finite element (FE) analysis. Subsection \ref{BVP} talks about the boundary value problem (BVP) and FE analysis while subsection \ref{Data} discusses the sample space and training and testing datasets. Subsection \ref{DNN} scrutinizes the architecture of the CNN model including the different layers and model parameters whilst subsection \ref{CNNeval} states the loss function and metrics used to evaluate the performance of the developed CNN model. Subsection \ref{GA} discusses the genetic algorithm used to find the optimal composite configurations. 

\subsection{BVP and FE analysis} \label{BVP}

The composite of interest is a 2D checkerboard system with two materials, soft and stiff. Both materials are assumed to be linear elastic and isotropic. The equilibrium equation in the absence of inertial and body forces is given by 
\begin{equation}\label{eqlbrm}
\begin{aligned}
    \sigma_{ij,j}=0\\
\end{aligned}
\end{equation}
where $\sigma_{ij}$ are the components of the Cauchy stress tensor while $()_{,j}$ is the divergence operator. Under the realm of linear elasticity, the constitutive relationship is expressed as
\begin{equation}\label{hooks}
\begin{aligned}
    \sigma_{ij}=E_{ijkl}\varepsilon_{kl}\\
    \end{aligned}
\end{equation}
where $\varepsilon_{kl}$ represent the components of an infinitesimal strain tensor, and $E_{ijkl}$ are the components of the fourth-order elasticity tensor. In case of three-dimensional analysis, $i, j, k, l=1,2,3$. Since small deformation is assumed, the strain is given by,
\begin{equation}\label{eps}
\begin{aligned}
    \varepsilon_{kl}=\frac{1}{2}(u_{k,l}+u_{l,k})\\
    \end{aligned}
\end{equation}
where $u_{k}$ are displacement components. $E_{ijkl}$ is defined by two constants when isotropic materials are assumed,
\begin{equation}
\begin{aligned}
    E_{ijkl}=\frac{E}{2(1+\nu)}(\delta_{ik}\delta_{jl}+\delta_{il}\delta_{jk})+\frac{3K\nu}{(1+\nu)}\delta_{ij}\delta_{kl}\\
    \end{aligned}
\end{equation}
where $E$ is Young's modulus, $\nu$ is Poisson's ratio, $K=\frac{E}{3(1-2\nu)}$ is the bulk modulus, and $\delta_{ij}$ is the Kronecker delta. Assuming plane stress condition reduces the governing equations to,
\begin{equation}
\begin{multlined}
\begin{aligned}
\sigma_{ij,j}=0 \quad i,j=1,2 \quad Equilibrium \; equation\\
\varepsilon_{kl}=\frac{1}{2}(u_{k,l}+u_{l,k}) \quad k,l=1,2 \quad Compatibility\\
\varepsilon_{11}=\frac{\sigma_{11}}{E}-\frac{\nu \sigma_{22}}{E}, \quad \varepsilon_{22}=\frac{\sigma_{22}}{E}-\frac{\nu \sigma_{11}}{E}, \quad \varepsilon_{12}=\frac{1+\nu}{E}\sigma_{12}.
\end{aligned}
\end{multlined}
\end{equation}

\begin{figure}[!htb]
    \centering
         \includegraphics[width=0.47\textwidth]{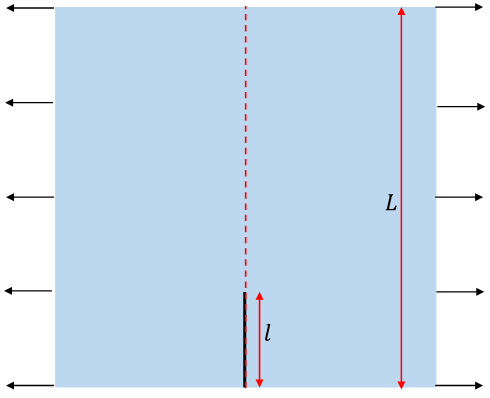}
          \caption{Illustration of the boundary value problem. The solid line shows the edge crack, $l/L=0.25$. The dashed line represents the plane of symmetry while the black arrows show the applied displacements.}
          \label{fracture}
     \end{figure}

After evaluating the quantities above, one can calculate $\varepsilon_{33}$ using $\varepsilon_{33}=\frac{-\nu}{E}(\sigma_{11}+\sigma_{22})$. In the present checkerboard composite system,  Young's modulus has a value of $E=1 \, GPa$ if the material is stiff and a value of $E=0.1 \, GPa$ if the material is soft. Poisson's ratio for both stiff and soft materials is set to be $\nu=\frac{1}{3}$. The failure strain of the stiff and brittle material is 10\% while the failure strain of the soft and ductile material is assumed to be 100\%. The composite system we are considering has a preexisting edge crack of 25\% of the specimen length in the y-direction. Figure \ref{fracture} depicts the crack, applied displacement boundary conditions, and plane of symmetry. Since symmetry is assumed, half of the problem is needed to be solved. We calculate three effective properties for the cracked composite system: modulus, strength, and toughness. Modulus is defined as the slope of the stress-strain curve of the cracked composite system while the strength is the maximum stress achieved by the system. The toughness is defined as the area under the stress-strain curve of the cracked composite system. In other words, it is defined as the energy needed to initiate the propagation of the crack. Classically these properties are defined in the literature when there is no preexisting crack. These quantities are defined here differently

\begin{figure}[!htb]
    \centering
         \includegraphics[width=0.6\textwidth]{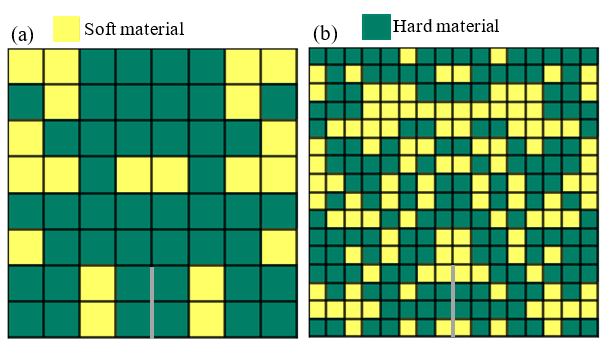}
          \caption{Examples of the randomly generated composite configurations: An example of the a) 8x8 grid and b) 16x16 grid.}
          \label{Comp_gen}
     \end{figure}

Finite element analysis (FEA) is done to evaluate the properties of interest. The FE simulation is stopped when the von Mises strain ($\varepsilon_{vM}$) at the element at the crack tip reaches the failure strain of the corresponding material. The von Mises strain is defined as
\begin{equation}
\begin{aligned}
\varepsilon_{vM}=\frac{2}{3}\Big(\frac{3}{2}(\varepsilon_{11}^2+\varepsilon_{22}^2+\varepsilon_{33}^2)+\frac{3}{4}\varepsilon_{12}^2\Big)^\frac{1}{2}.
\end{aligned}
\end{equation}
Four-node elements are used in the FEA where each element has four quadrature points, and each node has two degrees of freedom. Two different sizes of the cracked composite system are adopted: $8\times8$ and $16\times16$ system sizes. Figure \ref{Comp_gen} portrays examples of the randomly generated configurations. The finite element analysis is carried out using an in-house developed MATLAB code operated in the parallel computing toolbox for efficient data generation purposes. The code is validated by comparing the results obtained with those obtained from the open-source code FEAP \cite{taylor2014feap}.

\subsection{Data description and processing} \label{Data}
\begin{figure}[!htb]
  \makebox[\textwidth][c]{\includegraphics[width=1.5\textwidth]{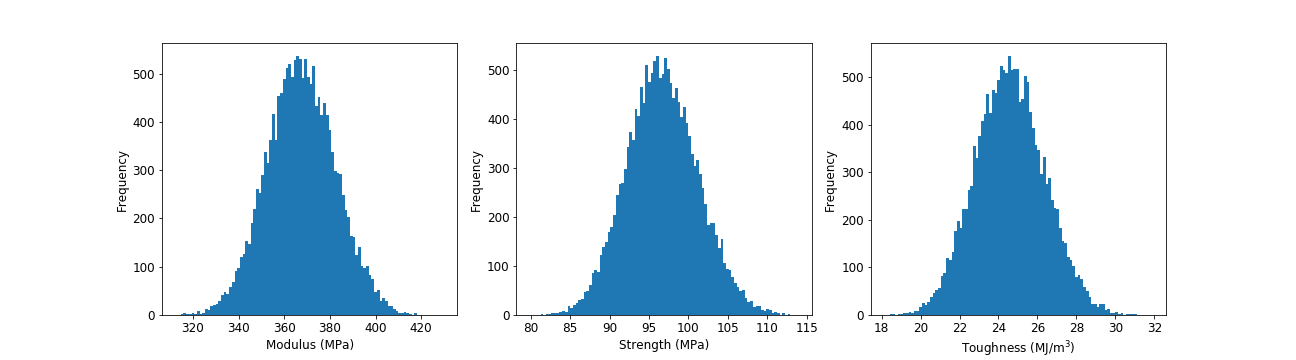}}%
  \caption{Histogram of the means obtained by repeated sampling for the 8x8 composite material.}
  \label{8x4Means}
\end{figure}

\begin{figure}[!htb]
  \makebox[\textwidth][c]{\includegraphics[width=1.5\textwidth]{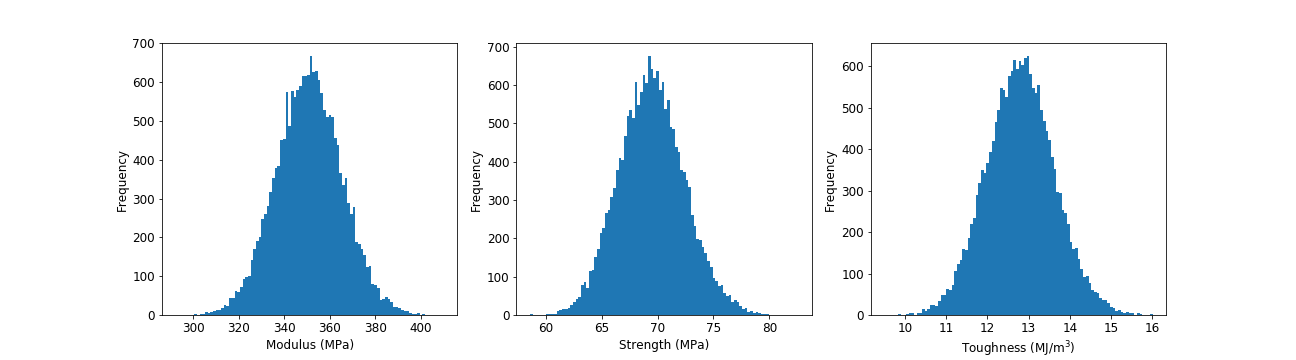}}%
  \caption{Histogram of the means obtained by repeated sampling for the 16x16 composite material.}
  \label{16x8Means}
\end{figure}

Finite element analysis is performed to generate the required data for this study. We consider two grid sizes, 8x8 and 16x16 elements. A large number of possible composite configurations corresponding to both grid sizes are examined by the FE method, and the resulting properties (modulus, strength, and toughness) of each configuration are stored. Configuration here refers to the microstructure which includes information about the volume fractions and spatial distributions of the phases. The sample space for the 8x8 grid with two-materials composites has a size of $2^{32}$ composite configurations while the 16x16 grid results in a sample space with a size of $2^{128}$ possible configurations. Due to the vastness of the sample space size, we rely on the central limit theorem for sampling our data. The sampling is performed in a batch-by-batch manner where all batches have the same fixed size where each batch is drawn from the sample space using a uniform distribution. Based on the central limit theorem, as the number of batches becomes larger, the distribution of means evaluated from repeated sampling converges to a normal distribution. The normal distribution is obtained regardless of the original distribution of the data as shown in Figures \ref{8x4Means} and \ref{16x8Means}. We end up with 4.3 million (M) data points for the 8x8 grid and 4.9M for the 16x16 grid. Tables \ref{table8x4} and \ref{table16x8} illustrate some key statistical information regarding the sampled data distributions. A perfect normal distribution has a skew and kurtosis of zero. The skews for the data corresponding to the 8x8 grid are 0.035, 0.112, and 0.085 for the modulus, strength, and toughness, respectively. The kurtoses for the data corresponding to the same grid are -0.041, -0.018, and -0.063 for the modulus, strength, and toughness, respectively. On the other hand, the skews for the data corresponding to the 16x16 grid are 0.031, 0.138, and 0.054 for the modulus, strength, and toughness, respectively. The kurtoses for the data corresponding to the same grid are 0.017, 0.035, and 0.060 for the modulus, strength, and toughness, respectively. 
\begin{table}[]
\centering
\caption{8x8 grid statistics}
\label{table8x4}
\resizebox{0.47\textwidth}{!}{%
\begin{tabular}{|l|l|l|l|}
\hline
Statistical parameter & Modulus & Strength & Toughness \\ \hline
Number of points & 4.3M & 4.3M & 4.3M \\ \hline
Coefficient of variation & 0.041 & 0.047 & 0.072 \\ \hline
Skew & 0.035 & 0.112 & 0.085 \\ \hline
Kurtosis & -0.041 & -0.018& -0.063\\ \hline
\end{tabular}%
}
\end{table}

\begin{table}[]
\centering
\caption{16x16 grid statistics}
\label{table16x8}
\resizebox{0.47\textwidth}{!}{%
\begin{tabular}{|l|l|l|l|}
\hline
Statistical parameter & Modulus & Strength & Toughness \\ \hline
Number of points & 4.9M & 4.9M & 4.9M \\ \hline
Coefficient of variation & 0.041 & 0.043 & 0.065 \\ \hline
Skew & 0.031 & 0.138 & 0.054 \\ \hline
Kurtosis & 0.017 & 0.035 & 0.06\\ \hline
\end{tabular}%
}
\end{table}

Furthermore, Monte Carlo analysis was performed to understand the data better and estimate the means (modulus, strength, and toughness means) of the entire sample spaces (populations) for the 8x8 and 16x16 grids. Figures \ref{8x4MC-Means} and \ref{16x8MC-Means} below illustrate the convergence of the population means for the 8x8 and 16x16 grids, respectively. It can be observed that after some number of batches the mean of the values stops fluctuating and stabilizes around a specific value, and this indicates that we have a good representation of the sample space. 

\begin{figure}[!htb]
  \makebox[\textwidth][c]{\includegraphics[width=1.5\textwidth]{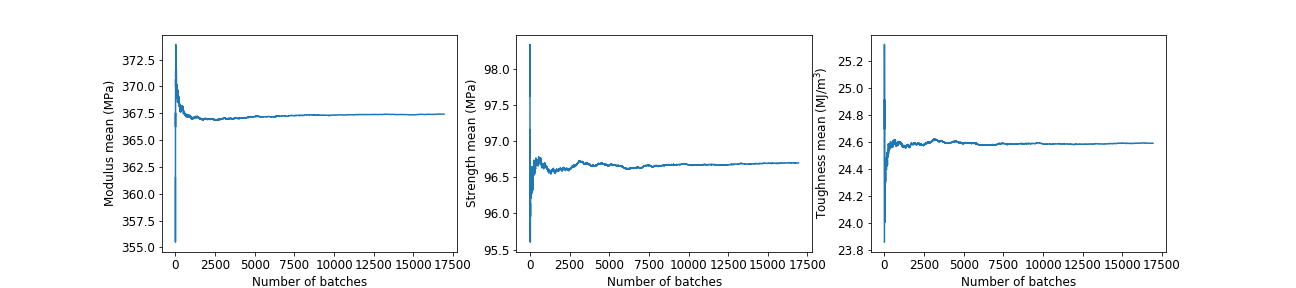}}%
  \caption{Monte Carlo analysis of the population means: 8x8 grid.}
  \label{8x4MC-Means}
\end{figure}

\begin{figure}[!htb]
  \makebox[\textwidth][c]{\includegraphics[width=1.5\textwidth]{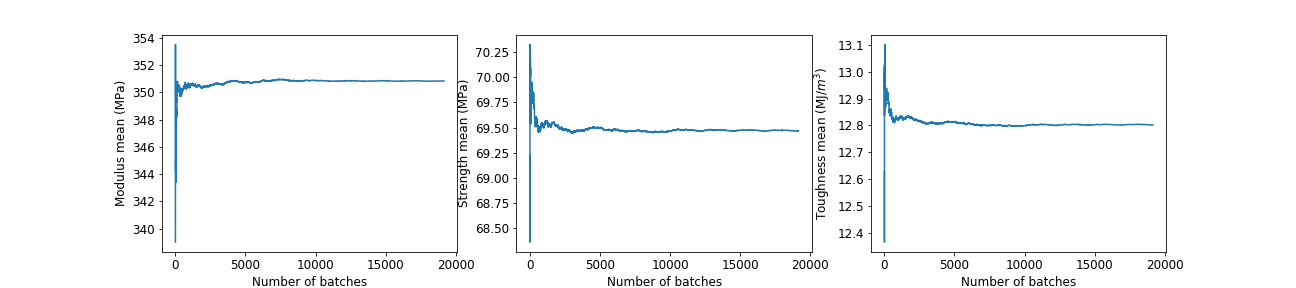}}%
  \caption{Monte Carlo analysis of the population means: 16x16 grid.}
  \label{16x8MC-Means}
\end{figure}

After acquiring the data from the FEA, some preprocessing steps are required to prepare the data to be fed into the CNN model. A matrix represents the data in its raw form (as acquired from the FEA). The rows of that matrix are the training examples (distinct configurations), and the columns represent the binary distribution of the materials (0 is stiff, and 1 is soft). On the other hand, each training example has a corresponding label vector containing the three material properties: modulus, strength, and toughness.

For our CNN model, the microstructure (two-material distribution) is defined as the feature (x) while the properties (modulus, strength, and toughness) of that specific composite are the label (y) to that feature. To be fed into the CNN model, each feature is transformed into an image form representing the spatial distribution of each material (8x8 and 16x16 grids) on which convolutions can then be performed to train the CNN model.

\subsection{Convolutional neural network model} \label{DNN}
Convolutional neural network is a widely-used class of deep neural networks that is remarkably successful in image recognition tasks. It has been widely used and incorporated in a lot of different disciplines. The main idea of a convolutional neural network is the presumption that the input data are images or can be interpreted as images. This helps in reducing the number of parameters significantly and thus results in faster processing.  

A typical convolutional neural network usually consists of the following layers: convolutional layers, pooling layers, fully connected layers, and activation functions. In the convolutional layer, a filter is applied to the input image through a convolution operation which preserves the spatial relationships between pixels in the image. On the other hand, a pooling layer applies a combining operation to the input such as a maximum pooling layer which outputs the maximum value element in each window. Fully connected layers and activation functions are similar to those used in simple DNN architectures. Additionally, other layers can be added to the CNN such as dropout layers which are a method of regularization used to reduce model overfitting  \cite{srivastava2014dropout}.

\begin{figure}[!htb]
    \centering
         \includegraphics[width=1\textwidth]{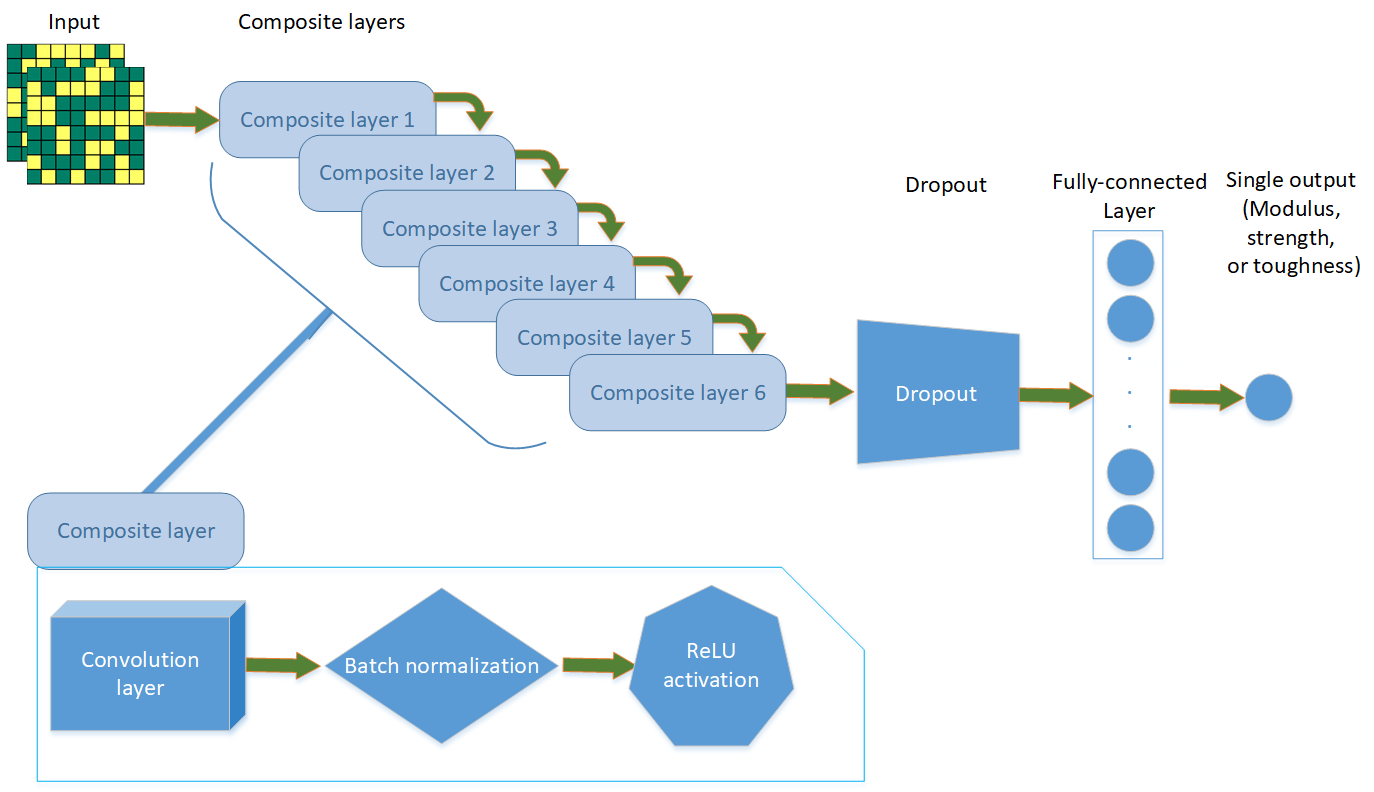}
          \caption{Illustration of the CNN model used in the present study.}
          \label{CNNArch}
     \end{figure}

We developed two CNN architectures, one for each grid size, where both architectures have six composite layers, a dropout layer, and a fully connected layer. Each composite layer comprises a 2D convolutional layer, a 2D batch normalization, and a rectified linear unit (ReLU) activation function. Kernel sizes of the convolutional layers are different in the two architectures. The same hyper-parameters are used for the two models as shown in Table \ref{default-hp}. Additionally, we use mini-batching to help the CNN models escape local minima and increase the convergence rate \cite{hinton2012practical}. Figure \ref{CNNArch} illustrates the architecture of the developed convolutional neural network.

\begin{table}[]
\centering
\caption{Hyper-parameter values used to design the models.}
\label{default-hp}
\resizebox{0.35\textwidth}{!}{%
\begin{tabular}{|l|l|}
\hline
Hyper-parameter & Value \\ \hline
Number of epochs & 200 \\ \hline
Batch size & 128 \\ \hline
Learning rate & 0.001 \\ \hline
Optimizer & Adam \\ \hline
Loss function & Mean square error \\ \hline
\end{tabular}%
}
\end{table}
     
\subsection{CNN model evaluation}\label{CNNeval}
We develop and test our models using PyTorch framework \cite{paszke2017pytorch}. Moreover, we use the Nvidia Pascal Titan XP GPU platform. In the training phase, we use the mean squared error cost function (MSE) to minimize the residual error between the model output and ground-truth data. The final model, which results after the training phase, uses the mean absolute percentage error (MAPE) to test the accuracy of the model when tested against new unseen data. Both the mean squared and mean absolute percentage errors are calculated as a discrepancy measure between the actual and predicted values using Equations \ref{RMSE} and \ref{MAPE},

\begin{equation}
\label{RMSE}
MSE = \Large{\frac{1}{n}\Sigma_{i=1}^{n}{({y_i -\hat{y_i}})^2}}
\normalsize
\end{equation}

\begin{equation}
\label{MAPE}
MAPE = \LARGE{\frac{1}{n}\sum_{i=1}^{n}|\frac{y_i-\hat{y_i}}{y_i}|}
\normalsize
\end{equation}
where $\hat{y_i}$ is the prediction obtained from the model, and $y_i$ is the actual (ground-truth) value. 

\subsection{Optimization using genetic algorithm}\label{GA}
Genetic algorithm (GA) is a metaheuristic inspired by Darwin's theory of evolution, and it belongs to the larger class of evolutionary algorithms. GA optimizers are often utilized to obtain optimal solutions relying on several bio-inspired operators, namely selection, crossover, and mutation \cite{mitchell1996introduction,holland1992adaptation,holland1975adaptation}. At each optimization step, the GA optimizer selects random (using the selection operator) individuals from the current generation to be parents producing children (offspring) for the next generation. The selection operator used here is based on the Roulette wheel mechanism where higher fitness values yield higher selection probabilities. Fitness is the value of the objective function in the optimization problem. The process of producing the offspring is done through the use of the crossover operator where the adopted crossover operator has two crossover points. On the other hand, the mutation operator applies random alterations to individual parents to form the children. The problem discussed here is binary-encoded since the genes have a value of either zero or one. Hence, mutated genes with an original value of zero end up with a value of one, and vice versa.

\begin{figure}[!htb]
    \centering
         \includegraphics[width=0.6\textwidth]{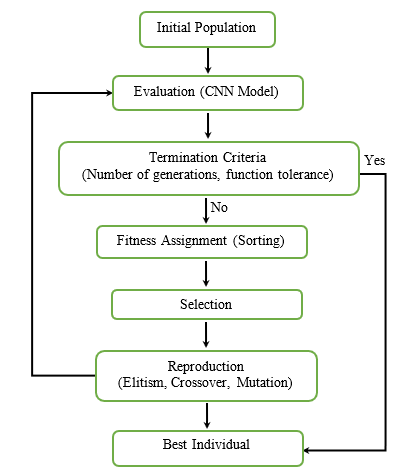}
          \caption{Procedure for the genetic algorithm used.}
          \label{GA_Alg}
     \end{figure}

In this paper, we adopted the elitism operator in addition to the main operators mentioned earlier where the elitism operator ensures that the chromosomes with best fitness values carry over to the next generation regardless of being a parent or child. This operator is very crucial as it guarantees that the solution quality attained by the optimizer is not decreased from one generation to the next \cite{baluja1995removing}. In the context of the present paper, a chromosome refers to a specific microstructure while a gene has the information about the material type at a particular element in the microstructure. The initial generation is selected randomly using a uniform distribution. The fitness evaluation of a microstructure (chromosome) is independent of the rest of the generation chromosomes. Thus, a generation can be evaluated easily in a parallel fashion. The procedure for the genetic algorithm used is portrayed in Figure \ref{GA_Alg}.

\section{Results and discussion}
\subsection{Linear model} \label{Linear}
We start by developing a linear model and use it as a baseline; the performance of the linear model is compared with the CNN model. The linear model developed here is used to quantitatively predict the properties of checkerboard composites, unlike the linear model developed by Gu et al. \cite{gu2018novo}. The linear model takes the general form shown in Equation 9,

\begin{equation}\label{LinearModel}
\begin{aligned}
Y = AX+B,
\end{aligned}
\end{equation}
where Y is the material property in question, X is the composite material distribution in vector form, and A and B are the weights. The weights (A and B) are calibrated through the training process to fit each material property as accurately as possible.

Since we have two composite grids (8x8 and 16x16), and we calibrate each model for the three material properties, we end up with 6 linear models. For the 8x8 grid, the 4.3M data points are used. The resulting coefficient of determination ($R^{2}$) for the fitted line is 0.916. However, the mean absolute percentage errors portray the shortcomings and oversimplification of the problem by the linear model where mean absolute percentage errors exceeded 25\% for the modulus and are as high as 40\% and 200\% for the strength and toughness, respectively. As for the 16x16 grid, the linear model results in a coefficient of determination ($R^{2}$) of 0.928, but the MAPE values remain too high with 24\%, 32\%, and 127\% for the modulus, strength, and toughness, respectively. The high error values produced by the linear models further justify the choice to utilize convolutional neural networks for our prediction problem.

Furthermore, the linear model is used to qualitatively determine the spatial location of the most critical elements in the composite. The essential elements are defined as the elements that have the highest weights attached to them in the linear model, and thus changes to them have the highest impact on the properties of the resulting composites. Figures \ref{linear8x4} and \ref{linear16x8} below show the ranking of the weights extracted from the linear model. The weights are ranked in descending order, and the positive weights are assigned positive ranks and vice versa. Positive weights correspond to the soft material while negative weights correspond to the stiff material as shown from the distribution of the weights. This is shown in Figures \ref{linear8x4} and \ref{linear16x8} where the highest ranked weights are near the crack tip, and the one directly at the crack tip is positive (soft material) and next to it are negative weights (stiff materials), which is in line with the physics of this problem which shows that a soft material placed right at the crack tip followed by stiff materials next to it yields best results in terms of material properties. Although the linear model fails to quantitatively predict the mechanical properties of checkerboard composites with high accuracy, one can still conclude some qualitative aspects as discussed in the present study and work of Gu et al. \cite{gu2018novo}.

\begin{figure}[!ht]
  \makebox[\textwidth][c]{\includegraphics[width=1\textwidth]{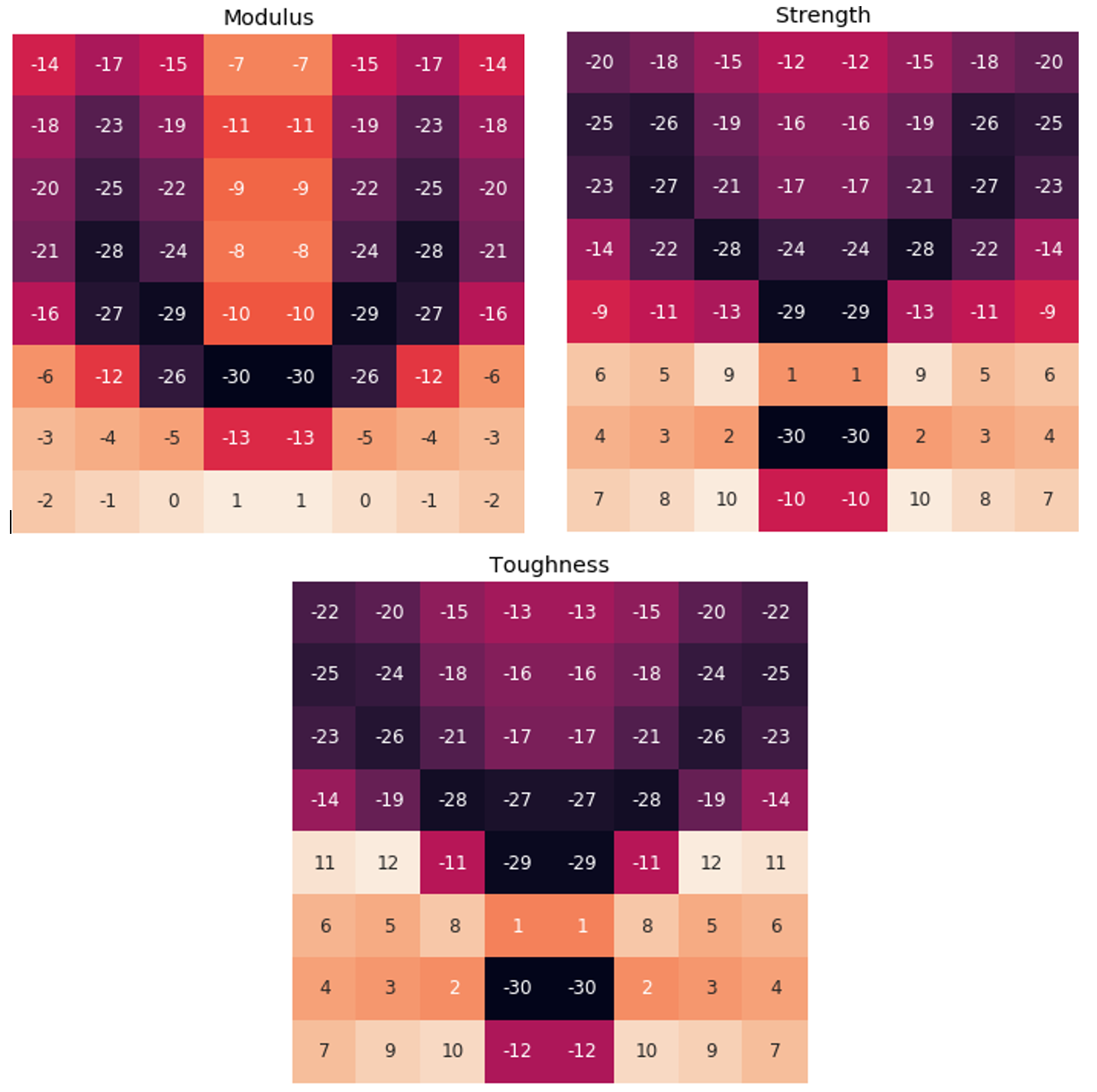}}%
  \caption{Weights ranking: 8x8 grid.}
  \label{linear8x4}
\end{figure}

\begin{figure}[!ht]
  \makebox[\textwidth][c]{\includegraphics[width=\textwidth]{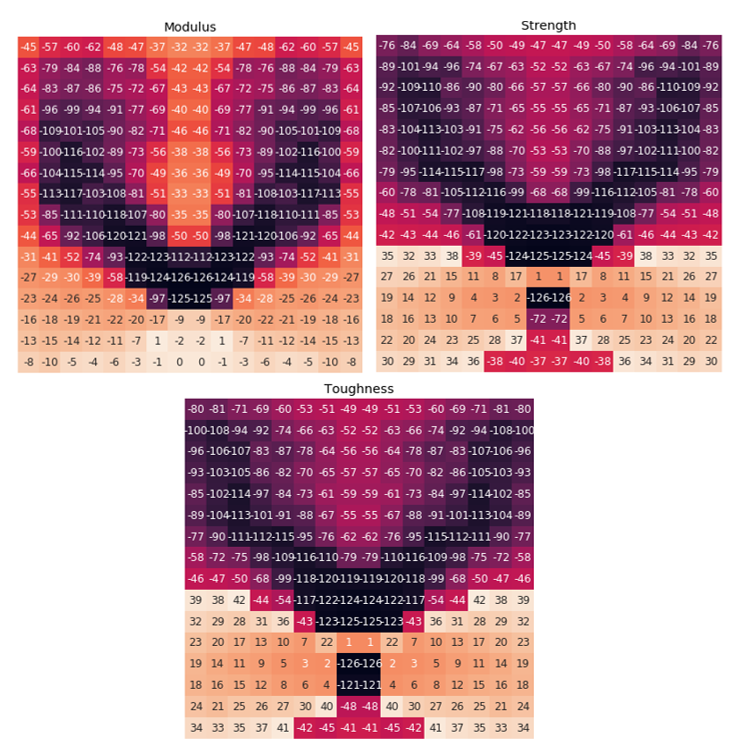}}%
  \caption{Weights ranking: 16x16 grid.}
  \label{linear16x8}
\end{figure}

\subsection{Convolutional neural network model} \label{CNN}

\begin{figure}[!ht]
    \centering
         \includegraphics[width=1\textwidth]{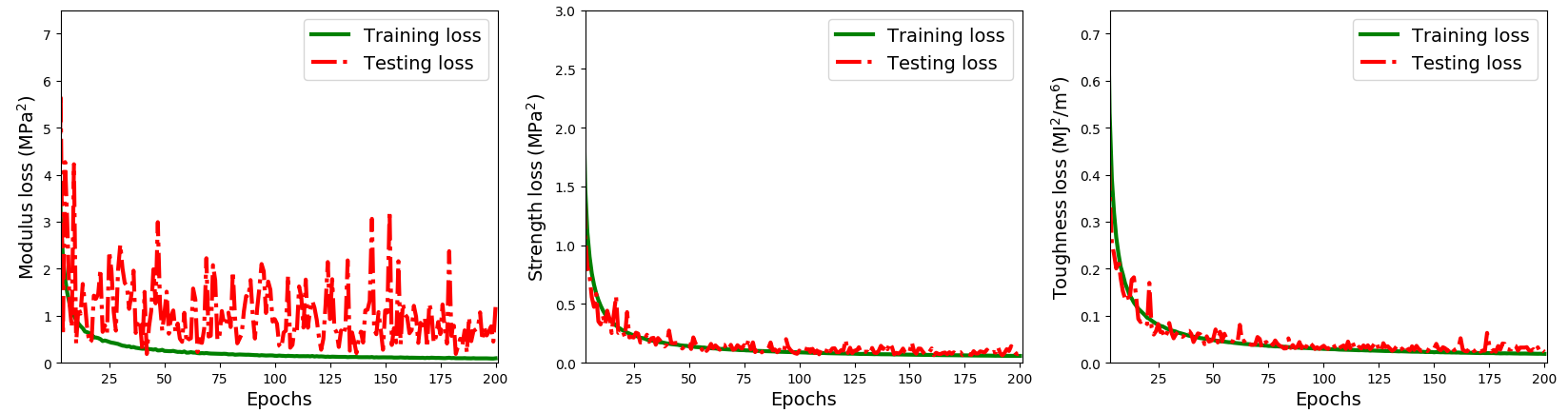}
          \caption{The convergence history of loss functions: 8x8 grid.}
          \label{loss8x4}
     \end{figure}

\begin{figure}[!ht]
\centering
\includegraphics[width=1\textwidth]{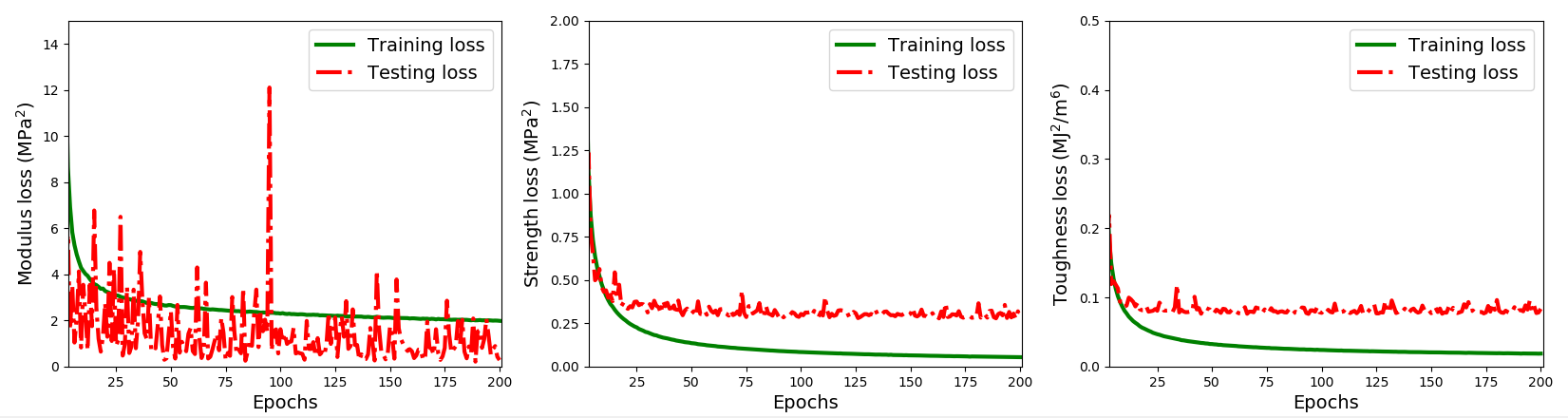}
\caption{The convergence history of loss functions: 16x16 grid.}
\label{loss16x8}
\end{figure}

CNNs have been intensively used in image recognition and signal processing, and they can be utilized to extract features from datasets. However, CNN models have to be trained before the inference process. The training process is an optimization problem in which the loss function (MSE) is minimized through the selection of the optimum weights of the CNN model. To start the training process, a dataset with many examples is required, so an optimal mapping approximating relationship between output and input data is determined. Here, we discuss the results obtained from the CNN model mentioned in subsection \ref{DNN} and summarized in Figure \ref{CNNArch}. The CNN model is trained using 200 epochs where an epoch is an iteration in the training process of the CNN model; one epoch is concluded when the CNN model is trained on every training example in the training dataset. The use of 200 epochs means that the CNN model encountered each training example in the training dataset 200 times. Figures \ref{loss8x4} and \ref{loss16x8} depict the convergence history of the training and testing loss functions for the case of the 8x8 grid and 16x16 grid, respectively. From the displayed results, one can see that the MSE loss functions of all properties for the two grid sizes converge to very low values. Also, the difference between the training and testing losses after the 200 epochs is tiny which in turn indicates that there is no critical overfitting occurring.  

The developed CNN model yields outstanding results for the defined problem, and it can predict the three material properties with very high accuracy. After the training process, we evaluate the performance of the CNN model using three parameters, the mean absolute percentage error, maximum error, and percentage of data points with an error greater than 5\%. These parameters are calculated using the testing dataset which is not seen by the CNN model throughout the training process of the model. The performance of the CNN model is summarized in Tables \ref{per8x4} and \ref{per16x8}. For both grid sizes, the MAPE of the modulus, strength, and toughness are less than 5\%. A stricter parameter to evaluate the performance of the CNN model is the maximum error. The maximum errors for the modulus and strength are less than 5\% while the maximum error for the toughness is larger than 5\%. However, the number of data points that have an error larger than 5\% is meager, 1.7\% in the case of the 8x8 grid and 0.082\% for the 16x16 grid. The results show that the CNN model outperforms the linear model discussed in subsection \ref{Linear}.

The excellent agreement between the results from the developed CNN model and ground-truth finite element results shows that machine learning algorithms in general and CNN models, in particular, have a high potential to be used in materials analyses and optimization. A possible extension to the current study is to build CNN models that capture the response of larger mechanical systems (e.g., 3D materials) and/or response of nonlinear materials. This leads us to another promising potentials of CNN models. Such CNN models can be integrated with optimization algorithms to find optimum solutions targeting various engineering applications. More details about optimization using CNN models are discussed in subsection  \ref{GAresults}.   

\begin{table}[]
\centering
\caption{Performance of the CNN model: 8x8 grid.}
\label{per8x4}
\begin{tabular}{l|l|l|l|}
\cline{2-4}
                                & MAPE  & Max error & \begin{tabular}[c]{@{}l@{}}\% of data points\\  with error \textgreater 5\%\end{tabular} \\ \hline
\multicolumn{1}{|l|}{Modulus}   & 0.6\% & 0.7\%     & 0.0\%                                                                                      \\ \hline
\multicolumn{1}{|l|}{Strength}  & 0.2\% & 0.4\%     & 0.0\%                                                                                      \\ \hline
\multicolumn{1}{|l|}{Toughness} & 0.8\% & 38.8\%    & 1.7\%                                                                                    \\ \hline
\end{tabular}
\end{table}

\begin{table}[]
\centering
\caption{Performance of the CNN model: 16x16 grid.}
\label{per16x8}
\begin{tabular}{l|l|l|l|}
\cline{2-4}
                                & MAPE  & Max error & \begin{tabular}[c]{@{}l@{}}\% of data points\\  with error \textgreater 5\%\end{tabular} \\ \hline
\multicolumn{1}{|l|}{Modulus}   & 0.2\% & 2.4\%     & 0.0\%                                                                                      \\ \hline
\multicolumn{1}{|l|}{Strength}  & 0.4\% & 2.1\%     & 0.0\%                                                                                      \\ \hline
\multicolumn{1}{|l|}{Toughness} & 2.4\% & 188.7\%   & 0.082\%                                                                                     \\ \hline
\end{tabular}
\end{table}

\subsection{Optimization using genetic algorithm}\label{GAresults}

Gradient-based topology optimization comprises several algorithms such as optimality condition \cite{khan1979optimality}, sequential linear programming \cite{lamberti2000comparison}, and sequential quadratic programming \cite{sedaghati2005benchmark}. Gradient-based optimization algorithms have fast convergence rates when used to search for an optimal solution. However, the sensitivity analyses of the objective functions and constraints are very challenging, especially if nonlinear materials are included \cite{james2014failure,james2015topology}. Also, most of the topology optimization problems are nonconvex. Hence, the selection of the starting point is a critical step in the optimization process which in turn makes it not easy. 

To overcome these issues, gradient-free optimization algorithms can be adopted. Generally, gradient-free optimization algorithms require many function evaluations when compared to gradient-based optimization algorithms. However, these function evaluations can be performed using parallel programming because each one is independent of the rest. Function evaluations may take a significant amount of time if the problem we are interested in solving is a nonlinear finite element problem. Therefore, developing CNN models that are capable of predicting the performance of materials and then integrating it with a gradient-free optimizer accelerate the optimization process significantly as the evaluation using a CNN model is way faster than the evaluation obtained from FE analysis \cite{do2019material}.

Here, we integrate the CNN model discussed earlier with a GA optimizer to find the optimum composite configurations yielding maximum mechanical properties: modulus, strength, and toughness. Genetic algorithms are a compromise between strong and weak search methods \cite{golberg1989genetic}. Strong methods progress with the search in an informed manner by using gradient information while weak methods (e.g., random and exhaustive search) perform the search in an uninformed manner through the extensive sampling of the design space \cite{chapman1994genetic}. In contrast, GA operates with (a) a strong progression toward designs with optimal performance and (b) weak operations of probabilistic pairing, crossover, and mutation. Generally, GA progresses toward regions in the design space with optimal performance without getting stuck in local optima (maxima/minima) in case of multimodal design space, a design space with multiple local optima.

Any successful optimization algorithm should have the right balance between the exploration and exploitation where exploration is related to global search (search throughout the design space for regions with good solutions), and exploitation is related to local search (solution refinement with the avoidance of big jumps) \cite{chen2009optimal}. Going too far with exploitation yields solutions with local optima, not necessarily global ones, while going too far with exploration results in a very slow convergence rate and a tremendous number of function evaluations. In GA, one can balance between exploration and exploitation through the selection of the different parameters such as crossover probability, mutation probability, and elitism ratio. 

In this paper, we discuss the optimization problem for the case of the 16x16 grid since the 8x8 grid is relatively simple. Hence, for the 16x16 grid, we have 128 genes (optimization parameters) after applying the symmetry boundary condition. Table \ref{GAparam} summarizes the parameters we adopted in solving the optimization problem. The parameters reported in Table \ref{GAparam} are used after tuning them based on a parametric study investigating the effect of different parameters on the optimal solutions obtained and convergence rate.

\begin{table}[]
\centering
\caption{Parameters used in the optimization process.}
\label{GAparam}
\begin{tabular}{|l|l|}
\hline
Generation size               & 1024  \\ \hline
Number of genes               & 128   \\ \hline
Maximum number of generations & 150   \\ \hline
Probability of crossover      & 0.95  \\ \hline
Number of crossover points    & 2     \\ \hline
Probability of mutation       & 0.005 \\ \hline
Elitism ratio                 & 0.10  \\ \hline
\end{tabular}
\end{table}

\begin{figure}[!ht]
  \makebox[\textwidth][c]{\includegraphics[width=1\textwidth]{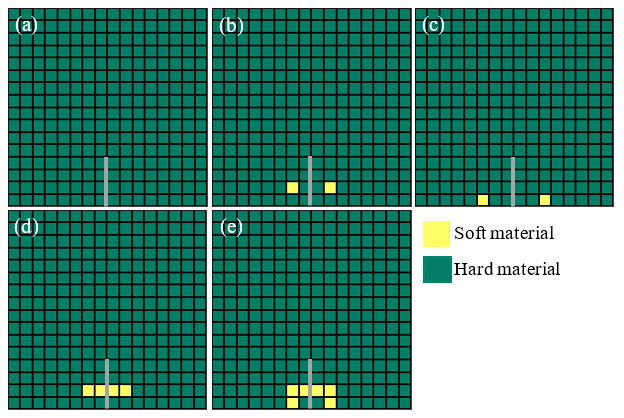}}%
  \caption{Genetic algorithm results: Modulus optimization. The volume fractions of the soft materials are: a) 0.0\%, b) 0.78\%, c) 0.78\%, d) 1.56\%, and e) 2.34\%.}
  \label{GAM}
\end{figure}

\begin{figure}[!ht]
  \makebox[\textwidth][c]{\includegraphics[width=1\textwidth]{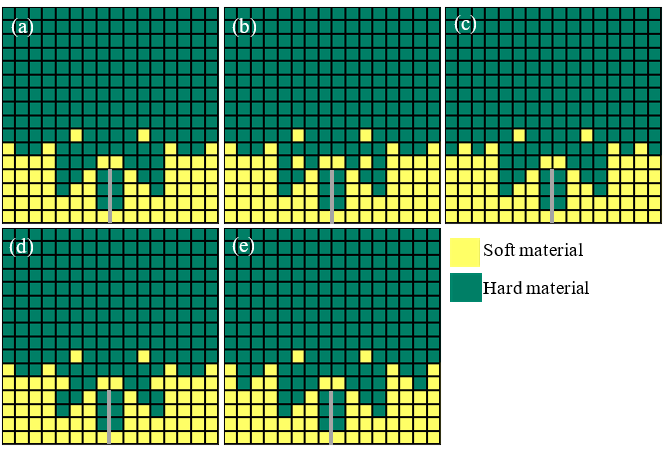}}%
  \caption{Genetic algorithm results: Strength optimization. The volume fractions of the soft materials are: a) 26.6\%, b) 27.3\%, c) 26.6\%, d) 27.3\%, and e) 25.8\%.}
  \label{GAS}
\end{figure}

\begin{figure}[!ht]
  \makebox[\textwidth][c]{\includegraphics[width=1\textwidth]{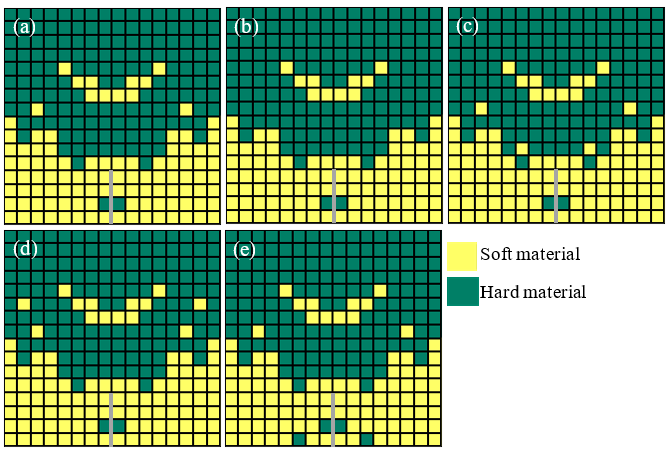}}%
  \caption{Genetic algorithm results: Toughness optimization. The volume fractions of the soft materials are: a) 40.6\%, b) 39.8\%, c) 41.4\%, d) 41.4\%, and e) 39.8\%.}
  \label{GAT}
\end{figure}

Figures \ref{GAM}-\ref{GAT} portray the top five composite configurations obtained from the GA optimizer when we optimize for modulus, strength, and toughness, respectively. The stopping criteria of the optimization problems are: (1) there is no significant change in the fitness function and (2) the maximum number of generations is met. Although the GA optimizer does not guarantee that we obtain the global optimum, it progresses toward the global optimum and approaches it with an appropriate tolerance. For the case of the modulus (see Figure \ref{GAM}) one concludes that the global optimum is met. From physics, the highest modulus is obtained when the composite is entirely composed of the stiff material. The configurations yielding top five modulus values have a volume fraction ranging from 0.00\% to 2.34\% of the soft material where the volume fraction of 0.00\% of the soft material corresponds to the global maximum. 

On the other hand, it is not straightforward to predict the optimum composite configuration for the case of strength and toughness. However, it is intuitive that we need a soft material in the region near the crack tip, and this is what Figures \ref{GAS} and \ref{GAT} are showing. Although the volume fractions of the phases are fixed in the work of Gu et al. \cite{gu2018novo}, similar conclusion, the need for soft elements next to the crack tip, is obtained. They did not use GA for the optimization; they introduced an optimization scheme based on the weights of the linear model. We discuss the drawbacks of the linear model in subsection \ref{Linear}. The configurations yielding top five strength values have a volume fraction ranging from 25.8\% to 27.3\% of the soft material. For the case of toughness, the volume fraction of the soft material ranges from 39.8\% to 41.4\%. 

Also, we have considered the case of multi-objective optimization. There are four possible combinations to simultaneously optimize for: a) modulus and strength, b) modulus and toughness, c) strength and toughness, d) modulus, strength, and toughness. The aggregate objective function (AOF), the function combining the different objectives into a scalar function, based on simple weighted sum of the objective functions performs poorly in case of non-convex Pareto frontier \cite{messac2015optimization}. One possible solution is to use compromise programming, 
\begin{equation}\label{MOO_AOF}
\begin{aligned}
AOF = w_m F_{m}^n + w_s F_{s}^n + w_t F_{t}^n,\\
F_{m} = \frac{modulus}{maximum \; modulus},\\
F_{s} = \frac{strength}{maximum \; strength},\\
F_{t} = \frac{toughness}{maximum \; toughness},
\end{aligned}
\end{equation}
where $n$ is the exponent of the objective functions, and it is selected to be $n=4$ in this study. $w_m$, $w_s$, and $w_t$ are the weights of modulus, strength, and toughness, respectively. In the case of optimizing for two properties, the weights corresponding to the two properties we are optimizing for are assigned a value of $0.5$ while the weight of the third property is assigned a value of $0.0$. In the case of optimizing for three properties, the three weights have a value of $0.333$. Also, the different objectives (modulus, strength, and toughness) have different scales which may cause some biases to objectives possessing higher values. This issue is resolved by normalizing each objective (property) by its maximum value (obtained from single-objective optimization (see Equation \ref{MOO_AOF}). 

\begin{figure}[!ht]
  \makebox[\textwidth][c]{\includegraphics[width=0.9\textwidth]{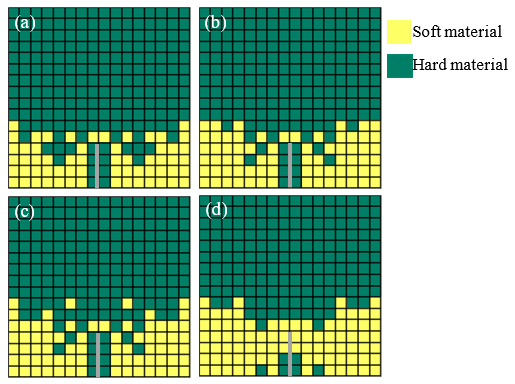}}%
  \caption{Multi-objective optimization using genetic algorithm: The volume fractions of the soft materials are: a) 23.4\% (modulus and strength), b) 27.3\% (modulus and toughness), c) 28.1\% (strength and toughness), and d) 32.8\% (modulus, strength, and toughness).}
  \label{MOO}
\end{figure}

Figure \ref{MOO} and Table \ref{MOO_results} show the results obtained from GA optimizer when multi-objective cases are considered. As concluded from the single-objective optimization, the modulus is maximum when there is no soft material. However, when we optimize for the strength and/or toughness in addition to the modulus, the optimizer tends to balance between the different objectives. Hence, the modulus obtained from the multi-objective optimization is significantly reduced due to the addition of soft materials, and the volume fraction of the soft material ranges from 23.4\% to 32.8\% depending on the properties we are optimizing. On the other hand, configurations with high values of strength and toughness have better harmony; optimizer tends to possess soft material in the region next to the crack tip. Consequently, less compromising is needed.   

\begin{table}[]
\centering
\caption{Results of the multi-objective optimization using GA.}
\label{MOO_results}
\begin{tabular}{|l|l|l|l|l|}
\hline
Output                                                                      & \begin{tabular}[c]{@{}l@{}}Modulus and\\  strength\end{tabular} & \begin{tabular}[c]{@{}l@{}}Modulus and\\  toughness\end{tabular} & \begin{tabular}[c]{@{}l@{}}Strength and\\  toughness\end{tabular} & \begin{tabular}[c]{@{}l@{}}Modulus, strength,\\  and toughness\end{tabular} \\ \hline
\begin{tabular}[c]{@{}l@{}}Volume fraction\\  of soft material\end{tabular} & 23.4\%                                                          & 27.3\%                                                           & 28.1\%                                                            & 32.8\%                                                                      \\ \hline
$F_m$                                                                       & 0.84                                                            & 0.81                                                             & 0.79                                                              & 0.76                                                                        \\ \hline
$F_s$                                                                       & 0.99                                                            & 0.99                                                             & 0.99                                                              & 0.97                                                                        \\ \hline
$F_t$                                                                       & 0.86                                                            & 0.94                                                             & 0.94                                                              & 0.97                                                                        \\ \hline
\end{tabular}
\end{table}
\section{Conclusions}
We develop a convolutional neural network model that is capable of quantitatively predicting the mechanical properties (modulus, strength, and toughness) of 2D checkerboard composites. The model is trained using finite element results (ground-truth data), and then it is tested on another dataset which is not seen by the model throughout the training process to ensure the validity of the model. The model shows very promising capabilities; it illustrates the potential of utilizing CNN models in structural and materials analysis. The developed CNN model is integrated with a genetic algorithm optimizer to obtain the composite configurations (material distribution and volume fraction) leading to materials with improved performance. CNN models have the potential of accelerating the current optimization techniques, and it might revolutionize the field of structural and materials design.
\section*{Acknowledgment}
This research was partially supported by the NSF Center for Novel High Voltage/Temperature Materials and Structures [NSF I/UCRC (IIP-1362146)]. R.A. would like to acknowledge the support of the Beckman Institute at the University of Illinois to carry out this work during his visiting scholar residence. The authors thank Dr. Grace X. Gu for helpful discussions.

\bibliography{mybibfile}

\end{document}